\documentclass[twoside]{article}
\usepackage[accepted]{aistats2019}
\usepackage[square,numbers,sort,compress]{natbib}

\usepackage{here}
\usepackage[utf8]{inputenc} % allow utf-8 input
\usepackage[T1]{fontenc}    % use 8-bit T1 fonts
\usepackage{hyperref}       % hyperlinks
\usepackage{url}            % simple URL typesetting
\usepackage{booktabs}       % professional-quality tables
\usepackage{amsfonts}       % blackboard math symbols
\usepackage{nicefrac}       % compact symbols for 1/2, etc.
\usepackage{microtype}      % microtypography
\usepackage{amsmath}
\usepackage{amssymb}
\usepackage{microtype}
\usepackage{graphicx}
\usepackage{subfigure}
\usepackage{booktabs} % for professional tables
\usepackage{bm}
\usepackage{amsthm}
\newtheorem{thm}{Theorem}
\newtheorem{lem}[thm]{Lemma}
\newtheorem{col}[thm]{Corollary}
\newtheorem{as}{Assumption}

\usepackage{color}

\begin{document}



\newcommand{\Com}[1]{\textcolor{blue}{#1}}
\newcommand*{\QEDA}{\hfill\ensuremath{\blacksquare}}%

\twocolumn[
\aistatstitle{Universal Statistics of Fisher Information in Deep Neural Networks: Mean Field Approach}

\aistatsauthor{Ryo Karakida \And Shotaro Akaho \And Shun-ichi Amari}
\aistatsaddress{AIST, Japan \And AIST, Japan \And RIKEN CBS, Japan} ]
%\aistatsauthor{ Author 1 \And Author 2 \And  Author 3 }
%\aistatsaddress{ Institution 1 \And  Institution 2 \And Institution 3 } 

\begin{abstract}
The Fisher information matrix (FIM) is a fundamental quantity to represent the characteristics of a stochastic model, including deep neural networks (DNNs). The present study reveals novel statistics of FIM that are universal among a wide class of DNNs. To this end, we use random weights and large width limits, which enables us to utilize mean field theories. We investigate the asymptotic statistics of the FIM’s eigenvalues and reveal that most of them are close to zero while the maximum eigenvalue takes a huge value. 
Because the landscape of the parameter space is defined by the FIM, it is locally flat in most dimensions, but strongly distorted in others.
Moreover, we demonstrate the potential usage of the derived statistics in learning strategies. First, small eigenvalues that induce flatness can be connected to a norm-based capacity measure of generalization ability. Second, the maximum eigenvalue that induces the distortion enables us to quantitatively estimate an appropriately sized learning rate for gradient methods to converge.
\end{abstract}

\section{Introduction}
\label{Introduction}
Deep learning has succeeded in  making hierarchical neural networks perform excellently in various practical applications \cite{nature2015}.  To proceed further, it would be beneficial to give more theoretical elucidation as to why and how deep neural networks (DNNs) work well in practice. 
In particular, it would be useful to not only clarify the individual models and phenomena but also explore various  unified theoretical frameworks that could be applied to a wide class of deep networks. One widely used approach for this purpose is to consider deep networks with random connectivity and a large width limit \cite{amari1974method,poole2016,schoenholz2016,pennington2017geometry,pennington2017nonlinear,pennington2018emergence,raghu2016expressive,daniely2016toward,li2018exploring,yang2017,xiao2018dynamical,kadmon2016,chen2018dynamical}. For instance, \citet{poole2016}  proposed a useful indicator to explain the expressivity of DNNs.
Regarding the trainability of DNNs, \citet{schoenholz2016}  extended this theory to backpropagation and found that the vanishing and explosive gradients obey a universal law.
These studies are powerful in the sense that they do not depend on particular model architectures, such as the number of layers or activation functions.

Unfortunately, such universal frameworks have not yet been established in many other topics. 
One is the geometric structure of the parameter space. 
For instance, the loss landscape without spurious local minima is
important for easier optimization and theoretically guaranteed in single-layer models \cite{mei2016landscape}, shallow piecewise linear ones \cite{soudry2017exponentially}, and extremely wide deep networks with the number of training samples  smaller than the width \cite{nguyen2017loss}.  
Flat global minima have been reported to be related to generalization ability through empirical experiments showing that  networks with such minima give better generalization performance \cite{hochreiter1997flat,keskar2016large}.  However,  theoretical analysis of the flat landscape has been limited in shallow rectified linear unit (ReLU) networks \cite{safran2016quality,tian2016symmetry}.
Thus,  a residual subject of interest is  to theoretically reveal the geometric structure of the parameter space truly common among various deep networks.

To establish the foundation of the universal perspective of the parameter space, 
this study analytically investigates the Fisher information matrix (FIM). As is overviewed in Section \ref{sec_2_2}, the FIM plays an essential role in the geometry of the parameter space and is a fundamental quantity in both statistics and machine learning.

\subsection{Main results}
This study analyzes the FIM of deep networks with random weights and biases, which are widely used settings to analyze the phenomena of DNNs \cite{amari1974method,poole2016,schoenholz2016,pennington2017geometry,pennington2017nonlinear,pennington2018emergence,raghu2016expressive,daniely2016toward,li2018exploring,yang2017,xiao2018dynamical,kadmon2016,chen2018dynamical}.  
First, we analytically obtain novel statistics of  the FIM, namely, the mean (Theorem \ref{thm_K0}), variance (Theorem \ref{thm_K1}), and  maximum of eigenvalues (Theorem \ref{thm4}). These are universal among a wide class of shallow and deep networks with various activation functions.
These quantities can be obtained from simple iterative computations of macroscopic variables.
To our surprise, the mean of the eigenvalues asymptotically decreases with an order of $O(1/M)$ in the limit of a large network width $M$, while the variance takes a value of $O(1)$, and the maximum eigenvalue takes a huge value of $O(M)$ by using the $O(\cdot)$ order notation.  
Since the eigenvalues are non-negative, these results mean that most of the eigenvalues are close to zero, but the edge of the eigenvalue distribution takes a huge value. 
Because the FIM defines the Riemannian metric of the parameter space,
the derived statistics imply that {\it the space is locally flat in most dimensions, but strongly distorted in others}. In addition, because the FIM also determines the local shape of a loss landscape,  
the landscape is also expected to be locally flat while strongly distorted.

Furthermore, to confirm the potential usage of the derived  statistics, we show some exercises. One is on the Fisher-Rao norm \cite{liang2017fisher} (Theorem \ref{thm_FR}).
This norm was originally proposed to connect the flatness of a parameter space to the capacity measure of generalization ability. We evaluate the Fisher-Rao norm by using an indicator of the small eigenvalues, $\kappa_1$ in Theorem \ref{thm_K0}. 
Another exercise is related to the more practical issue of determining the size of the learning rate necessary for the steepest descent gradient to converge. We demonstrate that  an indicator of the huge eigenvalue, $\kappa_2$ in Theorem  \ref{thm4}, enables us to {\it roughly estimate learning rates that make the gradient method converge to global minima} (Theorem \ref{thm7}). We expect that it will help to alleviate the dependence of learning rates on heuristic settings.  
%We also confirm the effectiveness of this estimation in numerical experiments.

\subsection{Related works}

Despite its importance  in statistics and machine learning,  study on the FIM for neural networks has been limited so far. This is because layer-by-layer nonlinear maps and huge parameter dimensions make it difficult to take analysis any further. Degeneracy of the eigenvalues of the FIM has been found in certain parameter regions \cite{fuk1996}.
To understand the loss landscape, \citet{pennington2017geometry} has utilized random matrix theory  and obtained the spectrum of FIM and Hessian under several assumptions, although the analysis is limited to special types of shallow networks.  
In contrast, this paper is the first attempt to apply the mean field approach, which overcomes the difficulties above  and enables us to identify universal properties of the FIM in various types of DNNs.

\citet{lecun1998efficient} investigated the Hessian of the loss, which coincides with the FIM at zero training error, and empirically reported that very large eigenvalues exist, i.e., ''big killers'', which affects the optimization (discussed in Section \ref{sec_6_1}). 
The eigenvalue distribution peaks around zero while its tail is very long; this behavior has been empirically known for decades \citep{sagun2017empirical}, but its theoretical evidence and evaluation have remained unsolved as far as we know. 
Therefore, our theory provides novel theoretical evidence that this skewed eigenvalue distribution and its huge maximum  appear universally in DNNs.   

The theoretical tool we use here is known as the {\it mean field theory} of deep networks  \cite{poole2016,schoenholz2016,li2018exploring,yang2017,xiao2018dynamical,kadmon2016,chen2018dynamical} as briefly overviewed in Section \ref{sec_MF_fin}. 
This method has been successful in analyzing neural networks with random weights under a large width limit and in explaining the performance of the models. In particular, it quantitatively coincides with experimental results very well and can predict appropriate initial values of parameters for avoiding the vanishing or explosive gradient problems \cite{schoenholz2016}. This analysis has been extended from fully connected deep networks to residual \cite{yang2017} and convolutional networks \cite{chen2018dynamical}. The evaluation of the FIM in this study is  also expected to be extended to such cases. 
%In Section 4.2, we will argue that the mean field theory is also beneficial to guess appropriate learning rates which prevents the gradient methods from exploding.  

\section{Preliminaries}
\subsection{Fisher information matrix (FIM)}
\label{sec_2_2}
We focus on the Fisher information matrix (FIM) of neural network models, which previous works have developed and is commonly used \cite{ama1998,amari2000adaptive,pascanu2013,ollivier2015riemannian,park2000,martens2015optimizing}. It is defined by  
\begin{equation}
F=\mathrm{E}[\nabla_\theta \log p(x,y;\theta) \nabla_\theta \log p(x,y;\theta)^T ],    
\end{equation}
where the statistical model is given by $p(x,y;\theta)=p(y|x;\theta)p(x)$. The output model is given by $p(y|x;\theta) = \exp(-||y-f_\theta(x)||^2/2)/\sqrt{2\pi}$, where $f_\theta(x)$ is the network output parameterized by $\theta$ and $||\cdot ||$ is the Euclidean norm.  The $q(x)$ is an input distribution. The expectation $\mathrm{E}[\cdot]$ is taken over the input-output pairs $(x,y)$ of the joint distribution $p(x,y;\theta)$. This FIM is transformed into  $F=\sum_{k=1}^C \mathrm{E}[ \nabla_\theta f_{\theta,k} (x) \nabla_\theta  f_{\theta,k} (x)^T]$, where $f_{\theta,k}$ is the $k$-th entry of the output ($k=1,...,C$). When $T$ training samples $x(t)$ $(t=1,...,T)$ are available,  the expectation can be replaced by the empirical mean. This is known as the empirical FIM and often appears in practice \cite{amari2000adaptive,pascanu2013,ollivier2015riemannian,park2000,martens2015optimizing}:
%\vspace{-5pt}
\begin{equation}
F = \frac{1}{T} \sum_{t=1}^T \sum_{k=1}^C \nabla_\theta f_{\theta,k} (t) \nabla_\theta  f_{\theta,k} (t)^T. \label{FIM}
\end{equation}
This study investigates the above empirical FIM for arbitrary $T$.
It converges to the expected FIM as $T \rightarrow \infty$.  Although the form of the FIM changes a bit in other statistical models (i.g., softmax outputs), these differences are basically limited to the multiplication of activations in the output layer \cite{park2000}. Our framework can be straightforwardly applied to such cases.

The FIM determines the asymptotic accuracy of the estimated parameters, as is known from a fundamental theorem of statistics, namely, the Cram\'er-Rao bound \cite{amari2016}. 
 Below, we summarize a more intuitive understanding of the FIM from geometric views.

\noindent
{\bf Information geometric view.}
Let us  define an infinitesimal squared distance $dr^2$, which represents the Kullback-Leibler divergence between the statistical model $p(x,y;\theta)$ and $p(x,y;\theta+d\theta)$ against a perturbation $d\theta$. It is given by 
\begin{equation}
 dr^2 := \mathrm{KL}(p(x,y;\theta)||p(x,y;\theta+d\theta))=d\theta^T F d\theta. 
\end{equation}
It means that the parameter space of a statistical model forms a Riemannian manifold and the FIM works as its Riemannian metric, as is known in information geometry \cite{amari2016information}. This quadratic form is equivalent to the robustness of a deep network:   
$\mathrm{E}[||f_{\theta+d\theta}(t) - f_\theta(t)||^2]=d\theta^T F d\theta. $
Insights from information geometry have led to the development of  natural gradient algorithms \cite{park2000,ollivier2015riemannian,martens2015optimizing} and, recently,  a capacity measure based on the  Fisher-Rao norm \cite{liang2017fisher}.

{\bf Loss landscape view.}
The empirical FIM (\ref{FIM}) determines the local landscape of the loss function around the global minimum.
Suppose we have a squared loss function $E(\theta)= (1/2T) \sum_t^T ||y(t)-f_\theta(t)||^2$.  The FIM is related to the Hessian of the loss function, $H := \nabla_{\theta} \nabla_{\theta} E(\theta)$,  in the  following way:
\begin{equation}
H = F - \frac{1}{T} \sum_t^T \sum_k^C (y_k(t)-f_{\theta,k}(t)) \nabla_{\theta} \nabla_{\theta} f_{\theta,k}(t). \label{eq_9}
\end{equation}
The Hessian coincides with the FIM when the parameter converges to the global minimum by learning,  that is, the true parameter $\theta^*$ from which the teacher signal $y(t)$ is generated by $y(t)=f_{\theta^*} (t)$ or, more generally, with noise (i.e., $y(t)=f_{\theta^*} (t)+\varepsilon_t$, where $\varepsilon_t$ denotes zero-mean Gaussian noise) \cite{amari2000adaptive}. 
In the literature on deep learning, 
its eigenvectors whose eigenvalues are close to zero locally compose flat minima, which leads to better generalization empirically  \cite{keskar2016large,liang2017fisher}. Modifying the loss function with  the FIM has also succeeded in overcoming the catastrophic forgetting \cite{kirkpatrick2017overcoming}.

Note that the information geometric view  tells us more than the loss landscape. While the Hessian (\ref{eq_9}) assumes the special teacher signal, the FIM works as the Riemannian metric to arbitrary teacher signals.

%Therefore, 

\subsection{Network architecture}
This study investigates a fully connected feedforward neural network. The network consists of one input layer with $M_0$ units, $L-1$ hidden layers ($L\geq 2$) with $M_l$ units per hidden layer $(l=1,2,...,L-1)$, and one output layer  with $M_L$ units:
\begin{equation}
u_i^{l}= \sum_{j=1}^{M_{l-1}} W_{ij}^{l} h_j^{l-1} +b_i^{l}, \ \ h_i^{l}= \phi(u_i^l). \label{eq1}
\end{equation}
This study focuses on the case of linear outputs, that is,  $f_{\theta,k}(x)=h^L_k=u_k^L$.
We assume that the activation function $\phi(x)$ and  its derivative $\phi'(x):=d\phi(x)/dx$ are square-integrable functions on a Gaussian measure.
A wide class of activation functions, including the sigmoid-like and (leaky-) ReLU functions, satisfy these conditions. Different layers may have different activation functions. 
Regarding the network width, we  set $M_l = \alpha_l M \ \ (l\leq L-1)$ and consider  the limiting case of large $M$ with constant coefficients  $\alpha_l$. 
This study mainly focuses on the case where the number of output units is given by a constant $M_L=C$. The higher-dimensional case of $C=O(M)$  is argued in Section \ref{sec3_4}.

The FIM (\ref{FIM}) of a deep network is computed by the chain rule in a manner similar to that of the backpropagation algorithm: 
\begin{align}
\frac{\partial f_{\theta,k}}{\partial W_{ij}^l} &= \delta_{k,i}^l \phi(u_j^{l-1}), 
\\   \delta_{k,i}^{l} &= \phi'(u_i^l) \sum_{j} \delta_{k,j}^{l+1} W_{ji}^{l+1},\ \  \delta^L_{k,k} = \phi'(u^L_k),
\label{b_chain}
\end{align}
where $\delta_{k,i}^{l} := \partial f_{\theta,k}/\partial u_i^l$ for ($k=1,...,C$). To avoid the complicated notation, we omit the index of the output unit, i.e.,  $\delta_{i}^{l} = \delta_{k,i}^{l}$,   in the following.

\subsection{Random connectivity}
The parameter set $\theta=\{W_{ij}^{l}, b_i^l \}$ is an ensemble generated by 
\begin{equation}
W_{ij}^l \sim \mathcal{N}(0,\sigma_{w^l}^2/M_{l-1}), \ \  
b_i^l \sim \mathcal{N}(0,\sigma_{b^l}^2), \label{eq2}
\end{equation}
and then fixed, where $\mathcal{N}(0,\sigma^2)$ denotes a Gaussian distribution with zero mean and variance $\sigma^2$, and we set  $\sigma_{w^l}>0$ and $\sigma_{b^l}>0$. 
To avoid complicated notation, we set them uniformly as $\sigma_{w^l}^{2} = \sigma_{w}^{2}$ and $\sigma_{b^l}^{2} = \sigma_{b}^{2}$, but they  can easily be generalized.  
It is essential to normalize the variance of the weights by $M$ in order to normalize the output $u_i^l$ to $O(1)$.  
 This setting is similar to how  parameters are initialized in practice \citep{glorot2010understanding}.
We also assume that the input samples $h_i^0(t) =x_i (t) \ \ (t=1,...,T)$ are generated in an i.i.d. manner from a standard Gaussian distribution:
$x_i(t) \sim \mathcal{N}(0,1).$ 
We focus here on the Gaussian case for simplicity, although we can easily generalize it to other distributions with finite variances. 

Let us remark that the above random connectivity is a common setting widely supposed in theories.
Analyzing such a network can be regarded as the typical evaluation \cite{amari1974method,poole2016,pennington2017geometry}.
It is also equal to analyzing the network randomly initialized \cite{safran2016quality,schoenholz2016}. 
The random connectivity is often assumed in the analysis of optimization as a true parameter of the networks, that is, the global minimum of the parameters \cite{tian2016symmetry,saad1995exact}.

\subsection{Mean-field approach}
\label{sec_MF_fin}
  On neural networks with random connectivity,  taking a  large width limit, we can  
 analyze the asymptotic behaviors of the networks. 
 Recently, this asymptotic analysis is referred to as the  mean field theory of deep networks, and we follow the previously reported notations and terminology \cite{poole2016,schoenholz2016,yang2017,xiao2018dynamical}.

  First,  let us introduce the following variables for feedforward signal propagations: $\hat{q}^l :=  \sum_i h^{l}_i(t)^2 /M_l$ and $ \hat{q}^l_{st} := \sum_{i}  h_i^{l} (s) h_{i}^{l}(t) /M_l$. 
  In the context of deep learning, these variables have been utilized to explain the depth to which signals can sufficiently propagate.
  The variable $\hat{q}^l_{st}$ is the correlation between the activations for different input samples $x(s)$ and $x(t)$ in the $l$-th layer. 
 Under the large $M$ limit,  these variables are given by integration over Gaussian distributions  
because the pre-activation $u_l^i$ is a weighted sum of independent random parameters and the central limit theorem is applicable  \citep{amari1974method,poole2016,schoenholz2016}:  
\begin{equation}
  \hat{q}^{l+1} = \int Du\phi^2 \left( \sqrt{q^{l+1}} u \right), \ \ 
q^{l+1} = \sigma_w^2 \hat{q}^l +\sigma_b^2,    \label{eq_hatq}  
\end{equation}
\begin{equation} 
\hat{q}^{l+1}_{st} =  I_{\phi}[q^{l+1},q_{st}^{l+1}],  \ \ 
q_{st}^{l+1} =\sigma_w^2 \hat{q}_{st}^l+\sigma_b^2, \label{eq_qst}
\end{equation}
with $\hat{q}^0 = 1$ and  $\hat{q}_{st}^0 = 0$ ($l=0, ..., L-1$). 
We can generalize the theory to unnormalized data with $\hat{q}^0 \neq 0$ and $\hat{q}^0_{st} \neq 0$, just by substituting them into the recurrence relations. 
The notation $Du = du \exp(-u^2/2) /\sqrt{2\pi}$ means integration over the standard Gaussian density. 
Here, the notation $I_\cdot[\cdot,\cdot]$ represents the following  integral: 
$ I_{\phi}[a,b] = \int Dz_1 Dz_2 \phi \left(\sqrt{a}z_1 \right) \phi \left(\sqrt{a} ( cz_1 + \sqrt{1-c^2} z_2) \right)$ with  $c= b/a$. 
%The derivation of these variables is simply based on the law of large number and the central limit theorem. 
%In the limit of $M \rightarrow \infty$, the reccurence relations are exact. 
The  $q_{st}^{l}$ is linked to the compositional kernel and utilized as the kernel of the Gaussian 
process \cite{lee2017deep}.

%The activations due to $x(s)$ and $x(t)$ are not independent, but rather have correlations because they share the same weight vectors even for different inputs.

Next, let us introduce variables for backpropagated signals: $\tilde{q}^l := \sum_i \delta_i^l(t)^2$ and  $\tilde{q}^l_{st} := \sum_{i}  \delta_i^{l} (s) \delta_{i}^{l}(t)$. Note that they are defined not by averages but by the sums. They remain $O(1)$ because of $C=O(1)$.
 $\tilde{q}^l_{st}$ is the correlation of backpropagated signals.
To compute these quantities, the previous studies assumed the following:
\begin{as}[\citet{schoenholz2016}]
{ \it On the evaluation of the variables $\tilde{q}^l$ and $\tilde{q}^l_{st}$,
one can use a different set of parameters, $\theta$ for the forward chain (\ref{eq1}) and $\theta’$ for the backpropagated chain (\ref{b_chain}), instead of using the same parameter set $\theta$ in both chains.
 }
\label{as1}
\end{as}
This assumption makes the dependence between  $\phi(u_i^l)$ (or $\phi'(u_i^l)$) and $\delta_j^{l+1}$, which share the same parameter set, very weak, and one can regard it as independent.
It enables us to apply the central limit theorem to the backpropagated chain (\ref{b_chain}).
Thus, the previous studies \cite{schoenholz2016,yang2017,xiao2018dynamical,pennington2018emergence} derived the following recurrence relations ($l=0,...,L-1$): 
\begin{equation}
 \tilde{q}^l = \sigma_w^2 \tilde{q}^{l+1} \int Du \left[\phi' (\sqrt{q^{l}}u) \right]^2,  \label{eq_tilq}
\end{equation}
\begin{equation}
\tilde{q}_{st}^l =  \sigma_w^2 \tilde{q}_{st}^{l+1}  I_{\phi'}[q^l,q_{st}^l],  \label{eq_tilqst} 
\end{equation}
 with $\tilde{q}^L= \tilde{q}_{st}^L=1$ because of the linear outputs.
 %$\tilde{q}^L = \int Du \left[\phi' (\sqrt{q^{L}}u) \right]^2$ and $\tilde{q}_{st}^L =I_{\phi'}[q^L,q_{st}^L]$.　
 The previous works confirmed excellent agreements  between the above equations and experiments. In this study, we also adopt the above assumption and use the recurrence relations.

The variables ($\hat{q}^l, \tilde{q}^l, \hat{q}_{st}^l, \tilde{q}_{st}^l$) depend only on the variance parameters $\sigma_w^2$ and $\sigma_b^2$, not on the unit indices. In that sense,  they are referred to as {\it macroscopic variables} (a.k.a. order parameters in statistical physics).  
The recurrence relations for the macroscopic variables simply require $L$ iterations of one- and two-dimensional numerical integrals.  Moreover, we can obtain their explicit forms for some activation functions (such as the error function, linear, and ReLU; see Supplementary Material B).

\section{Fundamental FIM statistics}
\label{sec_3}
Here, 
we report mathematical findings that  the mean,  variance, and maximum of eigenvalues of the FIM (\ref{FIM}) are explicitly expressed by using macroscopic variables.
 Our theorems are universal  for networks  ranging in size from shallow ($L=2$) to arbitrarily deep ($L \geq 3$) with various activation functions.

\subsection{Mean of eigenvalues}
\label{sec_3_1}
The FIM is a $P \times P$ matrix, where $P$ represents the total number of parameters. 
First, we compute the arithmetic mean of the FIM's eigenvalues as $ m_\lambda := \sum_{i=1}^P \lambda_i/P$. We find a hidden relation between the macroscopic variables and the statistics of FIM: 
\begin{thm}
{\it Suppose that Assumption 1 holds. In the limit of $M \gg 1$,
the mean of the FIM's eigenvalues is given by 
\begin{align}
m_{\lambda}  &=  C\frac{\kappa_1}{M}, \ \ \kappa_1:=  \sum_{l=1}^L\ \frac{\alpha_{l-1}}{\alpha} \tilde{q}^l  \hat{q}^{l-1}, \label{m_fin}
\end{align}
where $\alpha := \sum_{l=1}^{L-1} \alpha_l \alpha_{l-1}$.
The macroscopic variables $\hat{q}^l$ and $\tilde{q}^l$ can be computed recursively, and notably  $m_{\lambda}$ is $O(1/M)$.
}
\label{thm_K0}
\end{thm}
This is obtained from a relation $m_{\lambda}=\mathrm{Trace}(F)/P$ (detailed in  Supplementary Material A.1).
The coefficient $\kappa_1$ is a constant not depending on $M$, so it is $O(1)$. It is easily computed by $L$ iterations of the layer-wise recurrence relations (\ref{eq_hatq}) and (\ref{eq_tilq}). 

Because the FIM is a positive semi-definite matrix and its eigenvalues are non-negative, this theorem means that most of the eigenvalues asymptotically approach zero when $M$ is large. Recall that the FIM determines the local geometry of the parameter space. The theorem suggests that the network output remains almost unchanged against a perturbation of the parameters in many dimensions. It also suggests that the shape of the loss landscape is locally flat in most dimensions. 

Furthermore, by using Markov's inequality, we can  prove that the number of larger eigenvalues is limited, as follows: 
\begin{col}
{\it Let us denote the number of eigenvalues satisfying $\lambda\geq k$ by $N(\lambda \geq k)$ and suppose that Assumption 1 holds. For a constant $k>0$,  $N(\lambda \geq k) \leq \min \{\alpha  \kappa_1CM/k ,CT\}$ holds in the limit of $M \gg 1$. }
\label{thm_dim}
\end{col}
The proof is shown in Supplementary Material A.2. When $T$ is sufficiently small, we have a trivial upper bound $N(\lambda \geq k) \leq CT$ and the number of non-zero eigenvalue is limited. The corollary clarifies that even when $T$ becomes large, the number of eigenvalues whose values are $O(1)$ is $O(M)$ at most, 
and still much smaller than the total number of parameters $P$.

\subsection{Variance of eigenvalues}
\label{sec_3_2}

Next, let us consider the second moment $s_{\lambda} := \sum_{i=1}^P \lambda_i^2/P$. 
We  now  demonstrate that $s_{\lambda}$ can be computed from the macroscopic variables: 
\begin{thm}
{\it Suppose that Assumption 1 holds. In the limit of $M \gg 1$, the second moment of the FIM's eigenvalues  is  
\begin{align}
s_{\lambda} &= C \alpha \left(\frac{T-1}{T} \kappa_2^2 + \frac{1}{T}\kappa_1^2 \right), \\ 
\kappa_2 &:= \sum_{l=1}^L   \frac{\alpha_{l-1}}{\alpha} \tilde{q}^l_{st}\hat{q}^{l-1}_{st}.
\end{align}
The macroscopic variables
$ \hat{q}^l_{st}$ and   $\tilde{q}^l_{st}$
can be computed recursively, and $s_\lambda$ is $O(1)$.\footnote{Let us remark that we have assumed $\sigma_b>0$ in the setting (\ref{eq2}). If one considers a case of no bias term ($\sigma_b=0$), odd activations $\phi(x)$ lead to 
$\hat{q}_{st}^l=0$ and $\kappa_2=0$. In such exceptional cases,  we need to evaluate the lower order terms of $s_\lambda$ and $\lambda_{max}$ (outside the scope of this study).}
}
\label{thm_K1}
\end{thm}
The proof is shown in Supplementary Material A.3.

From Theorems 1 and 3, we can conclude that the variance of the eigenvalue distribution, $s_\lambda- m_{\lambda}^2$, is $O(1)$. Because the mean $m_\lambda$ is $O(1/M)$ and most eigenvalues are close to zero, this result means that the edge of the eigenvalue distribution takes a huge value.

\begin{figure}[t]
\vskip 0.2in
\begin{center}
\centerline{\includegraphics[width=1\columnwidth]{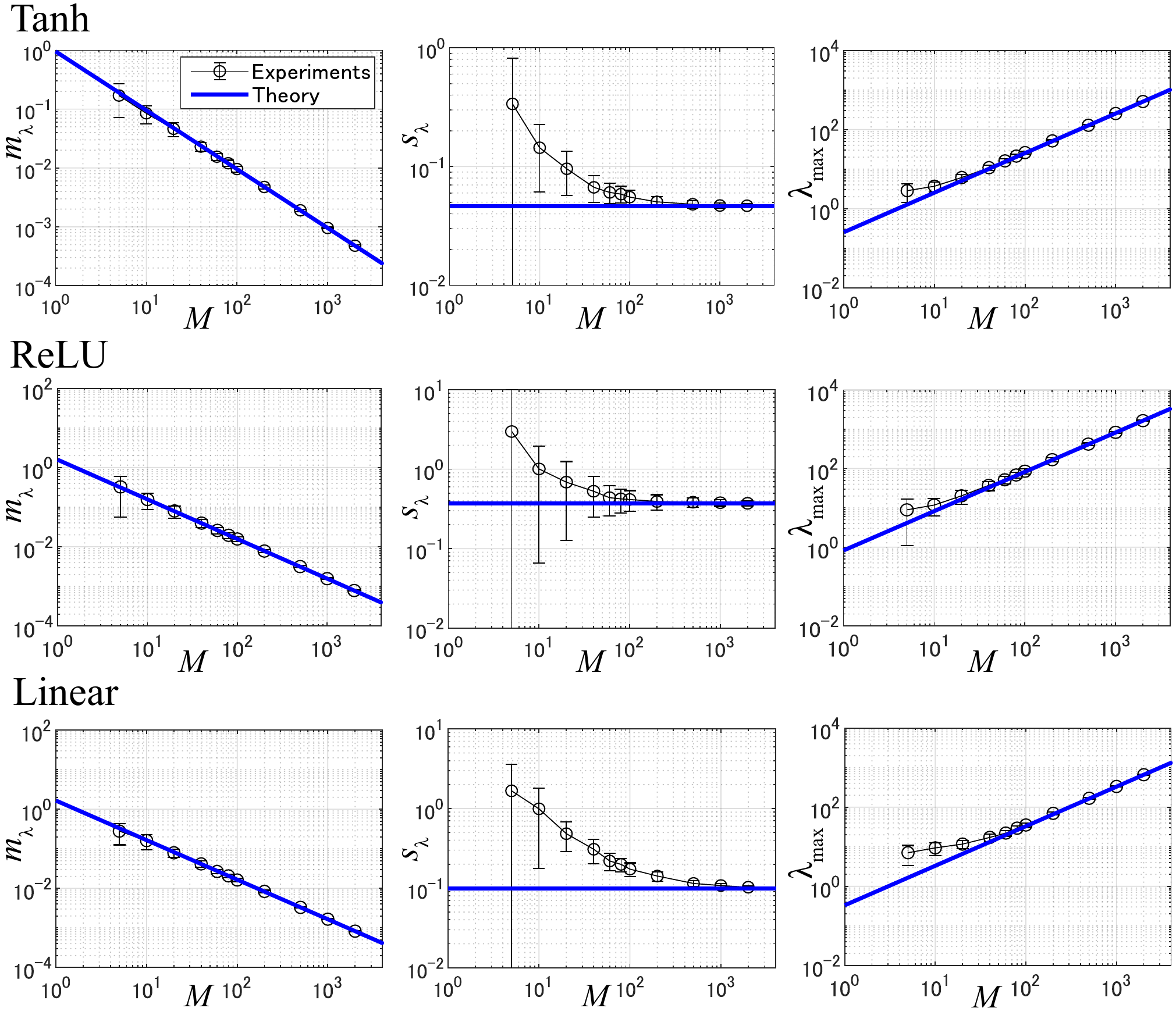}}
\caption{Statistics of FIM eigenvalues: means (left),  second moments (center), and maximum (right). Our theory predicts the results of numerical experiments, indicated by the black points and error bars. The experiments used 100 random ensembles with different seeds. The variances of the parameters were given by ($\sigma_w^2,\sigma_b^2$) = ($3,0.64$) in the tanh case, ($2,0.1$) in the ReLU case, and ($1,0.1$) in the linear case. Each colored line represents  theoretical results obtained in the limit of $M \gg 1$. 
}
%\label{fig1}
\end{center}
\vskip -0.2in
\end{figure}

\subsection{Maximum eigenvalue}
\label{sec_4_2}

As we have seen so far, the mean of the eigenvalues is $O(1/M)$, and the variance is $O(1)$. 
Therefore, we can expect that at least one of the eigenvalues must be huge. 
Actually, we can show that the maximum eigenvalue (that is, the spectral norm of the FIM) increases in the order of $O(M)$ as follows. 
\begin{thm}  \label{thm4}
{\it Suppose that Assumption 1 holds. In the limit of $M \gg 1$, the maximum eigenvalue of the FIM is }
\begin{equation}
\lambda_{max} = \alpha \left( \frac{T-1}{T} \kappa_2 + \frac{1}{T}\kappa_1 \right)M. \label{max}
\end{equation}
\end{thm}
The $\lambda_{max}$ is derived from the dual matrix $F^*$ (detailed in Supplemental Material A.4). 
If we take the limit $T \rightarrow \infty$, we can characterize the quantity $\kappa_2$ by the maximum eigenvalue as  $\lambda_{max} = \alpha \kappa_2 M$.
Note that $\lambda_{max}$ is independent of $C$.
When $C=O(M)$, it may depend on $C$, as shown in Section 3.4.

This theorem suggests that the network output changes dramatically  with a perturbation of the parameters in certain dimensions and that the local shape of the loss landscape is strongly distorted in that direction. Here, note that $\lambda_{max}$ is proportional to $\alpha$, which is the summation over $L$ terms. This means that,  when the network becomes deeper, the parameter space is more  strongly distorted.

We confirmed the agreement between our theory and numerical experiments, as shown in Fig. 1. Three types of deep networks with parameters generated by random connectivity (\ref{eq2}) were investigated: tanh, ReLU, and linear activations  ($L=3$, $\alpha_l=C=1$).  
The input samples were generated using i.i.d. Gaussian samples, and  $T=10^2$. 
When $P>T$, we calculated the eigenvalues by using the dual matrix $F^*$ (defined in Supplementary Material A.3) because $F^*$ is much smaller and its eigenvalues are easy to compute.
The theoretical values of $m_\lambda$, $s_\lambda$ and $\lambda_{max}$ agreed very well with the experimental values in the large $M$ limit. We could predict $m_{\lambda}$ even for small $M$. In addition,  
In Supplementary Material C.1, we also show the results of experiments with fixed $M$ and changing $T$. The theoretical values coincided with the experimental values very well for any $T$ as the theorems predict.

\section{Connections to learning strategies}
\label{sec_6}
%\vspace{-3pt}
Here, we show some applications that demonstrate how our universal theory on the FIM can potentially enrich deep learning theories.
It enables us to quantitatively measure the behaviors of learning strategies as follows.
%\vspace{-3pt}

\subsection{The Fisher-Rao norm}
\label{sec_6_2}

Recently, \citet{liang2017fisher} proposed the Fisher-Rao norm for a capacity measure of generalization ability:
\begin{equation}
||\theta||_{FR} = \theta^T F \theta,
\end{equation}
where $\theta$ represents weight parameters. 
They reported that this norm has several desirable properties  to explain the high generalization capability of DNNs. In deep linear networks,  its generalization capacity (Rademacher complexity) is upper bounded by the norm. In deep ReLU networks, the Fisher-Rao norm serves as a lower bound of the capacities induced by other norms, such as the path norm \cite{neyshabur2015norm} and the spectral norm \cite{specnorm}. 
 The Fisher-Rao norm is also motivated by information geometry, and  invariant under node-wise linear rescaling in ReLU networks. This is  a desirable property to connect capacity measures with flatness induced by the rescaling \cite{dinh2017sharp}.

Here, to obtain a typical evaluation of the norm, we define the average over possible parameters with fixed variances ($\sigma_w^2,\sigma_b^2$) by $\langle \cdot \rangle_\theta =  \int \prod_i D\theta_{i} (\cdot)$, which  leads to the following theorem: 
\begin{thm} \label{thm_FR}
{\it Suppose that Assumption 1 holds. In the limit of $M \gg 1$, the Fisher-Rao norm of DNNs satisfies 
%\vspace{-5pt}
\begin{equation}
\langle ||\theta||_{FR} \rangle_{\theta} \leq \sigma_w^2 \frac{\alpha}{\alpha_{min}} C\kappa_1, \label{FR_inq}
\end{equation}
where $\alpha_{min} = \min_i \alpha_i$. Equality holds in a network with a uniform width $M_l=M$, and then we have $\langle ||\theta||_{FR} \rangle_{\theta} = \sigma_w^2 (L-1)C \kappa_1$. 
} 
\end{thm}
%\vspace{-10pt}
The proof is shown in Supplementary Material A.6. 
Although what we can evaluate is only the average of the norm,  it can be quantified by $\kappa_1$. 
%In the supplementary material, we also show an explicit expression of the norm as computed by the macroscopic variables, that is, $\langle ||\theta||_{FR} \rangle_{\theta} = \sigma_w^2 C \sum_l \tilde{q}^l  \hat{q}^{l-1}$.
This guarantees that the norm is independent of the network width  in the limit of $M \gg 1$, which was empirically conjectured by \citep{liang2017fisher}.

Recently, \citet{smith2017understanding} argued that the Bayesian factor composed of the Hessian of the loss function, whose special case is the FIM, is related to the generalization. Similar analysis to the above theorem may enable us to quantitatively understand the relation between the statistics of the FIM and the indicators to measure the generalization ability.

\subsection{Learning rate for convergence}
\label{sec_6_1}
Consider the steepest gradient descent method in a batch regime.  Its update rule is given by  
\begin{equation}
\theta_{t+1} \leftarrow \theta_t  -\eta \frac{\partial E(\theta_t)}{ \partial \theta} + \mu(\theta_{t}-\theta_{t-1}), 
\label{eq20t}
\end{equation}
where $\eta$ is a constant learning rate. We have added a momentum term with a coefficient $\mu$ because it is widely used in training deep networks.  
Assume that the squared loss function $E(\theta)$ of Eq. (\ref{eq_9}) has a global minimum $\theta^*$ achieving the zero training error $ E(\theta^*) =0$.
Then, the FIM's maximum eigenvalue is dominant over the convergence of learning as follows:

\begin{lem}
{\it A learning rate satisfying $\eta <  2(1+\mu)/\lambda_{max}$ is necessary for the steepest gradient method to converge to the global minimum $\theta^*$. }
\label{Bottou}
\end{lem}
The proof is given by the expansion around the minimum, i.e., $E(\theta^*+d\theta)= d\theta^T F d\theta $ (detailed in Supplementary Material A.7). This lemma is a generalization of \citet{lecun1998efficient}, which proved the case of $\mu=0$. 
Let us refer to $ \eta_c := 2(1+\mu)/\lambda_{max}$ as the critical learning rate. When $\eta>\eta_c$, the gradient method never converges to the global minimum. 
The previous work \citep{lecun1998efficient} also claimed that  $\eta=\eta_c/2$ is the best choice for fastest convergence around the minimum.  Although we focus on the batch regime, the eigenvalues also determine the bound of the gradient norms and the convergence of learning  in the  online regime \citep{bottou1998online}. 

Then, combining Lemma \ref{Bottou} with Theorem \ref{thm4} leads to the following: 
\begin{thm}\label{thm7}
{\it Suppose that Assumption 1 holds. Let a global minimum $\theta^*$ be generated by Eq. (\ref{eq2}) and satisfying $ E(\theta^*) =0$. In the limit of $M \gg 1$, the gradient method never converges to  $\theta^*$ when}
%\vspace{-8pt}
\begin{equation}
 \eta > \eta_c, \ \  \eta_c := \frac{2(1+\mu)}{\alpha \left( \frac{T-1}{T} \kappa_2 + \frac{1}{T}\kappa_1 \right)M  }. \label{rate}
\end{equation}
\end{thm}
Theorem \ref{thm7} quantitatively reveals that, the wider the network becomes, the smaller the learning rate we need to set. 
In addition,  $\alpha$  is the sum over $L$ constant positive terms, so a deeper network requires a finer setting of the learning rate and it will make the optimization more difficult. 
In contrast, the expressive power of the network grows exponentially as the number of layers increases \citep{poole2016,montufar2014number}.  We thus expect  there to be a trade-off between trainability and expressive power. 

\begin{figure}[t!]
\vskip 0.2in
\begin{center}
\centerline{\includegraphics[width=1\columnwidth]{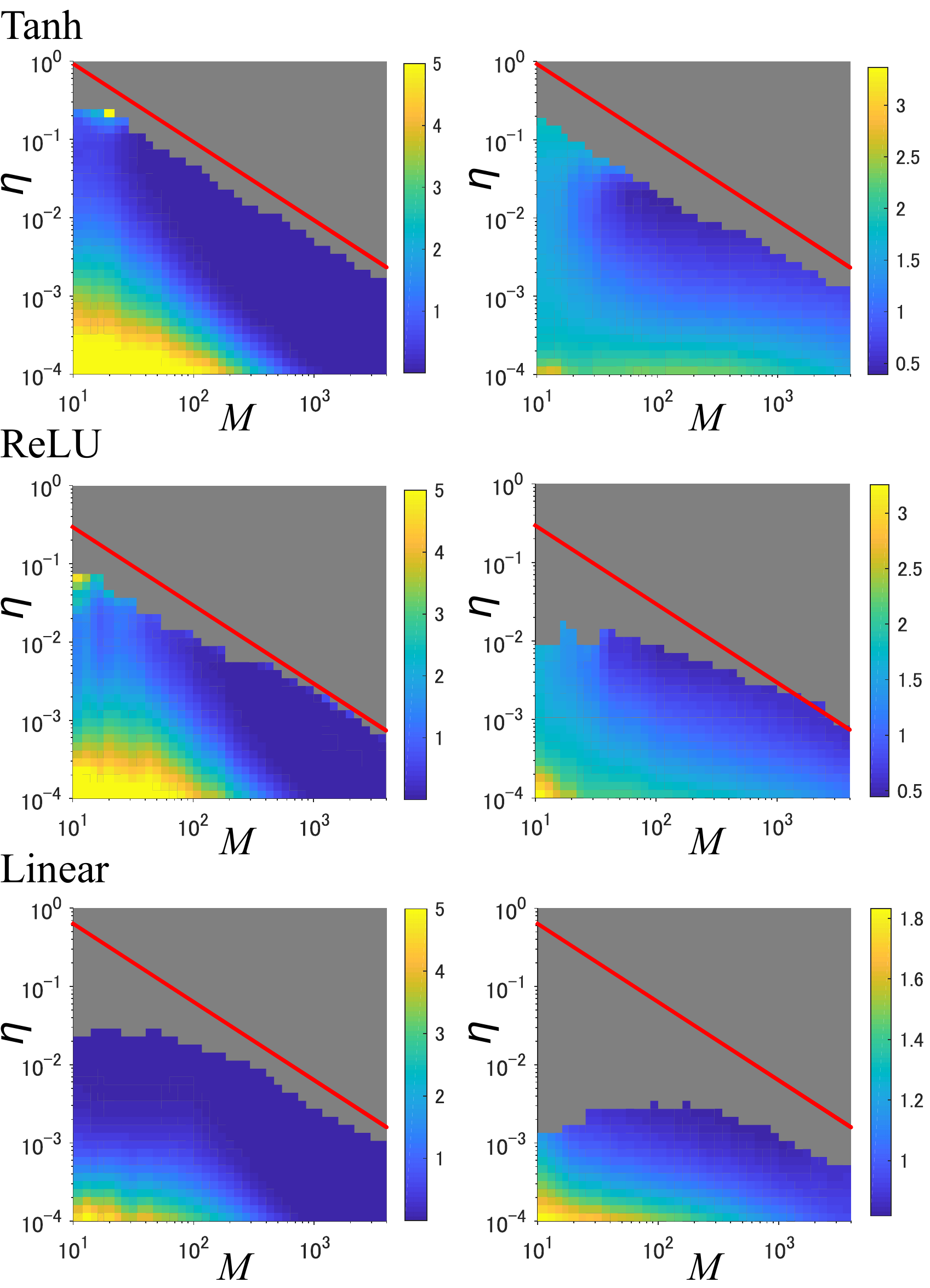}}
\caption{Color map of training losses: Batch training on artificial data (left column) and SGD training on MNIST (right column).  The losses are averages over five trials. The color bar shows the value of the training loss after the training. The region where the loss diverges (i.e., is larger than 1000) is in gray. The red line shows the theoretical value of $\eta_c$. The initial conditions of the parameters were taken from a Gaussian distribution (\ref{eq2}) with $(\sigma_w^2,\sigma_b^2)=(3,0.64)$ in tanh networks,  $(2,0.1)$ in ReLU networks, and $(1,0.1)$ in linear networks.}
%\label{fig1}
\end{center}
\vskip -0.2in
\end{figure}

To confirm the effectiveness of Theorem \ref{thm7},  we performed several experiments. As shown in Fig. 2, we exhaustively searched training losses while changing $M$ and $\eta$, and found that the theoretical estimation coincides well with the experimental results. We trained deep networks ($L=4$, $\alpha_l=1$, $C=10$)  and the loss function was given by the squared error.

The left column of Fig. 2 shows the results of training on
artificial data. We generated training samples $x(t)$ in the Gaussian manner ($T=100$) and  teacher signals $y(t)$ by the teacher network with a true parameter set $\theta^*$ satisfying Eq. (\ref{eq2}). 
We used the gradient method (\ref{eq20t}) with $\mu=0.9$ and trained the DNNs for $100$ steps. The variances $(\sigma_w^2,\sigma_b^2)$ of the initialization of the parameters were set to the same as the global minimum.  
We found that the losses of the experiments were clearly divided into two areas: one where the gradient exploded (gray area) and the other where it was converging (colored area).  
The red line is  $\eta_c$ theoretically calculated using $\kappa_1$ and $\kappa_2$ on $(\sigma_w^2,\sigma_b^2)$ of the initial parameters.
Training on the regions above $\eta_c$ exploded, just as Theorem \ref{thm7} predicts. The explosive region with $\eta<\eta_c$ got smaller in the limit of large $M$.

We performed similar experiments on benchmark datasets and found that the theory can estimate the appropriate learning rates. The results on MNIST are shown in the right column of Fig. 2. 
As shown in Supplementary Material C.2, the results of training on CIFAR-10 were almost the same as those of MINIST.
We used stochastic gradient descent (SGD) with a mini-batch size of $500$ and $\mu=0.9$, and trained the DNNs for $1$ epoch. 
Each training sample was $x(t)$ normalized to zero mean and variance $1$ ($T=50000$).  
The initial values of  $(\sigma_w^2,\sigma_b^2)$ were set to  the vicinity of the special parameter region, i.e., the critical line of the order-to-chaos transition,  which the previous works \citep{poole2016,schoenholz2016} recommended to use for achieving high expressive power and trainability. 
Note that the variances $(\sigma_w^2,\sigma_b^2)$ may change from the initialization to the global minimum, 
and the conditions of the global minimum in Theorem \ref{thm7} do not hold in general. Nevertheless, the learning rates estimated by Theorem \ref{thm7}  explained the experiments well. Therefore, the ideal conditions supposed in Theorem \ref{thm7} seem to  hold effectively. This may be explained by the conjecture that the change from the initialization to the global minima is  small in the large limit \cite{neyshabur2018towards}. 

Theoretical estimations of learning rates in deep 
networks have so far been limited; such gradients as AdaGrad and Adam also require heuristically determined hyper-parameters for learning rates.
Extending our framework would be  beneficial in guessing learning rates to prevent the gradient update from exploding.  

\subsection{Multi-label classification with high dimensionality}
\label{sec3_4}
This study mainly focuses on the multi-dimensional output of $C=O(1)$. This is because the number of labels is much  smaller than the number of hidden units in most practice cases. However, since classification problems with far more labels are sometimes examined in the context of machine learning \cite{deng2010does}, it would be helpful to remark on the case of $C=O(M)$ here. 
Denote the mean of the FIM's eigenvalues in the case of $C=O(M)$ as $m_\lambda'$ and so on. Straightforwardly, we can derive  
\begin{equation}
  m_\lambda'=m_\lambda, \ \  s_{\lambda} \leq s_{\lambda}' \leq Cs_{\lambda}, 
\end{equation}
\begin{equation}
\lambda_{max} \leq \lambda_{max}' \leq \sqrt{\alpha  Cs_{\lambda}}M.
\end{equation}
The derivation is shown in Supplementary Material A.5. 
The mean of eigenvalues has the same form as Eq. (\ref{m_fin}) obtained in the case of $C=O(1)$.  The second moment and maximum eigenvalues can be evaluated by the form of inequalities. We found that the mean is of $O(1)$ while the maximum eigenvalue is  of $O(M)$ at least and of $O(M^2)$ at most. 
Therefore, the eigenvalue distribution is more widely distributed than the case of $C=O(1)$.
%Because the large eigenvalues arethis may  

\section{Conclusion and discussion}

The present work elucidated the asymptotic statistics of the Fisher information matrix (FIM) common among deep networks with any number of layers and various activation functions.
The statistics of FIM are characterized by the small mean of eigenvalues and the huge maximum eigenvalue, which are  computed by the recurrence relations. This suggests that the parameter space determined by the FIM is locally flat in many directions while highly distorted in certain others. 
As examples of how one can connect the derived statistics to learning strategies,
  we suggest the Fisher-Rao norm and learning rates of steepest gradient descents. 
  
 We demonstrated that the experiments with the Gaussian prior on the parameters coincided well with the theory. Basically, the mean field theory is based on the central limit theorem with the parameters generated in an i.i.d. manner with finite variances. Therefore, one can expect that the good agreement with the theory is not limited to the experiments with the Gaussian prior. Further experiments will be helpful to clarify the applicable scope of the mean field approach.

The derived statistics are also of potential importance to other learning strategies, for instance, natural gradient methods. 
When the loss landscape is non-uniformly distorted, naive gradient methods are likely to diverge or become trapped in plateau regions, but the natural gradient, $F^{-1} \nabla_\theta E(\theta)$, converges more efficiently \citep{amari2000adaptive,park2000,pascanu2013,ollivier2015riemannian}. Because it normalizes the distortion of the loss landscape, 
the naive extension of Section \ref{sec_6_1} to the natural gradient leads to $\eta_c=2(1+\mu)$ and it seems to be much easier to choose the appropriately sized learning rate. However, 
we found that the FIM has many eigenvalues close to zero, and 
the inversion of it would make the gradient very unstable. In practice, 
several experiments showed that the choice of damping term $\varepsilon$, introduced in $ (F+\varepsilon I)^{-1} \nabla_\theta E(\theta)$, is crucial to its performance in DNNs  \citep{martens2015optimizing}. The development of practical natural gradient methods will require modification such as damping.

It would also be interesting for our framework to quantitatively reveal the effects of normalization methods on the FIM.  In particular, batch normalization may alleviate the larger eigenvalues because it empirically allows larger learning rates for convergence \citep{ioffe2015batch}.
It would  also be fruitful to investigate the eigenvalues of the Hessian with a large error  (\ref{eq_9}) and to theoretically quantify the negative eigenvalues that lead to the existence of saddle points and the loss landscapes without spurious local minima  \citep{sfn2014}. The global structure of the parameter space should be also explored.  
We can hypothesize that the parameters are globally connected through the locally flat dimensions and compose manifolds of flat minima. 

Our framework on FIMs is readily applicable to other architectures such as convolutional networks and residual networks by using the corresponding mean field theories \cite{yang2017,xiao2018dynamical}. 
 To this end, it may be helpful to remark that macroscopic variables in residual networks essentially diverge at the extreme depths \citep{yang2017}. 
 If one considers extremely deep residual networks,  the statistics  will require a careful examination of the order of the network width and the explosion of the macroscopic variables.  
We expect that further studies will establish a mathematical foundation of deep learning  from the perspective of the large limit.

\subsubsection*{Acknowledgments}
This work was partially supported by a Grant-in-Aid for Research Activity Start-up (17H07390) from the Japan Society for the Promotion of Science (JSPS).

\bibliographystyle{unsrtnat}
%\small{
\bibliography{test}
%}

\onecolumn

\part*{Supplementary Materials}
\appendix
\setcounter{section}{0}
\renewcommand{\thesection}{\Alph{section}}

\section{Proofs}

\section*{A.1 Theorem 1}
\label{A_thm1}

\renewcommand{\theequation}{A.\arabic{equation} }
\setcounter{equation}{0}
\subsection*{(i) Case of $C=1$}

To avoid complicating the notation, we first consider the case of the single output ($C=1$). The general case is shown after. 
The network output is denoted by $f(t)$ here.
We  denote the Fisher information matrix with full components as   
\begin{equation}
F= \sum_{t=1}^T\begin{bmatrix}   \nabla_W f(t) \nabla_W f(t)^T  &  \nabla_W f(t) \nabla_b  f(t)^T  \\ \nabla_b f(t) \nabla_W  f(t)^T  &  \nabla_b f(t) \nabla_b  f(t)^T  \end{bmatrix}/T,
\end{equation}
where  we notice that
\begin{equation}
\nabla_{b^l_i}  f(t) = \delta_i^l(t).
\end{equation}
In general, the sum over the eigenvalues is given by the matrix trace, $m_{\lambda} = \mathrm{Trace}(F)/P$. 
We denote the average of the eigenvalues of the diagonal block  as $m_{\lambda}^{(W)}$ for $ \nabla_W f \nabla_W  f^T$, and $m_{\lambda}^{(b)}$ for  $ \nabla_b f \nabla_b  f^T $. Accordingly, we find
\begin{equation}
m_{\lambda} = m_{\lambda}^{(W)} +  m_{\lambda}^{(b)}.
\end{equation} 

The contribution of $m_{\lambda}^{(b)}$ is negligible in the large $M$ limit as follows. 
The first term is
\begin{align}
m_{\lambda}^{(W)} &=  \sum_{t=1}^T \mathrm{Trace} (\nabla_W f(t) \nabla_W  f(t)^T)/(TP) \\
&=  \sum_{t=1}^T \sum_l \sum_{i,j} \delta^{l}_i(t)^2 h^{l-1}_j(t)^2 /(TP). \label{eq_mean}
\end{align}
We can apply the central limit theorem to summations over the units  $\sum_i \delta^{l}_i(t)^2$ and $\sum_j h^{l-1}_j(t)^2$  independently because they do not share the index of the summation.
By taking the limit of $M \gg 1$, we obtain
$\sum_i \delta^{l}_i(t)^2 \sum_j h^{l-1}_j(t)^2/M_{l-1} =\tilde{q}^l \hat{q}^{l-1}$. The variable $\hat{q}^{l-1}$ is computed by the recursive relation (9). Under the Assumption 1, $\tilde{q}^{l}$ is given by the  recursive relation (11). Note that this transformation to the macroscopic variables holds regardless of the sample index $t$. Therefore, we obtain
\begin{align}
m_{\lambda}^{(W)}  &=  \kappa_1/M, \ \ \kappa_1:= \sum_{l=1}^L\ \frac{\alpha_{l-1}}{\alpha} \tilde{q}^l  \hat{q}^{l-1},
\end{align}
where $\alpha_l$ comes from $M_l=\alpha_l M$, and $\alpha$ comes  from $P = \alpha M^2$. 
%Let us remark on a technical issue of the proof. 
%The $m_{\lambda}$ is a bit similar to the gradient of the backpropagation $\nabla_{W_{ij}} h$. To analytically evaluate the L2 norm of $\nabla_{W_{ij}} h$,  
%the previous work assumed that the backpropagated signal $\delta^{l+1}_i(t)$ is independent of the forward signal $h^{l}_j(t)$ ($l=1,2,...,L-1$) for any $t$. In contrast, what we proved is the summation over $\nabla_{W_{ij}} h$ ($i=1, ..., M_{l-1},j=1,...,M_l$). 

In contrast, the contributions of the bias entries are smaller than those of the weight entries in the limit of $M\gg 1$, as is easily confirmed: 
\newpage
\begin{align}
m_{\lambda}^{(b)} &=  \sum_t \mathrm{Trace} (\nabla_b f(t) \nabla_b  f(t)^T)/(TP) \\ 
&=\sum_t \sum_l \sum_{i} \delta_i^l(t)^2/(TP) \\ 
&= \sum_l\tilde{q}^l/(\alpha M^2) \ \ \ \ (\mathrm{when} \ \ M \gg 1).
\end{align}
$m_{\lambda}^{(W)}$ is  $O(1/M)$ while $m_{\lambda}^{(b)} $ is $O(1/M^2)$. Hence, the mean $m_{\lambda}^{(b)}$ is negligible and we obtain $m_{\lambda}=\kappa_1/M$.

\subsection*{(ii) $C>1$ of $O(1)$}
 We can apply the above computation of $C=1$ to  each network output $\nabla f_k $ ($k=1,...,C$):
\begin{equation}
\mathrm{Trace}(\nabla_\theta f_k\nabla_\theta f_k^T/T)/P = \kappa_1/M.
\end{equation}
Therefore, the mean of the eigenvalues becomes  
\begin{align}
m_{\lambda} &=\sum_k^C \mathrm{Trace}(\nabla_\theta f_k\nabla_\theta f_k^T/T)/P \\
&=C\kappa_1 /M.
\end{align}

\QEDA

\section*{A.2 Corollary 2}
Because the FIM is a positive semi-definite matrix, its eigenvalues are non-negative.
For a constant $k>0$, we obtain
\begin{align}
m_{\lambda} &= \frac{1}{P} \left (\sum_{i; \lambda_i<k} \lambda_i + \sum_{i; \lambda_i \geq k} \lambda_i \right ) \\ 
&\geq  \frac{1}{P} \sum_{i; \lambda_i \geq k} \lambda_i \\
& \geq \frac{1}{P}  N(\lambda \geq k) k.
\end{align}
This is known as Markov's inequality.   
%Because $\mathrm{P} (\lambda \geq a)$ multiplied by the number of all eigenvalues ($P$) is equal to $N(\lambda \geq a)$ by definition,
When $M \gg 1$, combining this with Theorem 1 immediately yields 
\begin{align}
N(\lambda \geq k) \leq \alpha  \kappa_1CM/k.  
\end{align}
Because $CT$ is also a trivial upper bound of $N(\lambda \geq k)$, we obtain Corollary 2.
%for $M \gg 1$ and $T \gg 1$.
\QEDA

\section*{A.3 Theorem 3}
Let us describe the outline of the proof.
One can express the FIM as $F=(BB^T)/T$  by definition. Here, let us consider a dual matrix of $F$, that is, $F^*:=(B^TB)/T$. $F$ and $F^*$ have the same nonzero eigenvalues. Because the sum of squared eigenvalues is equal to $\mathrm{Trace}(F^*(F^*)^T)$, we have $s_\lambda= \sum_{s,t}^T (F^*_{st})^2/P $. The non-diagonal entry $F^*_{st}$ $(s \neq t)$  corresponds to an inner product of the network activities for different inputs $x(s)$ and $x(t)$, that is, $\kappa_2$. 
The diagonal entry $F^*_{ss}$ is given by $\kappa_1$. 
Taking the summation of $(F^*_{st})^2$ over all of $s$ and $t$, we obtain the theorem. In particular, when $T=1$ and $C=1$, $F^*$ is equal to the squared norm of the derivative $\nabla_\theta f$, that is, $F^* = ||\nabla_{\theta} f||^2 $, and one can easily check $s_\lambda = \alpha \kappa_1^2$.

The detailed proof is given as follows.

\subsection*{(i) Case of $C=1$}

Here, let us express the FIM as $F= \nabla_\theta f \nabla_\theta f^T/T$, where $\nabla_\theta f$  is a $P\times T$ matrix whose columns are the gradients on each input sample, i.e.,  $\nabla_{\theta} f(t)$ $(t=1,...,T)$.
We also introduce a dual matrix of $F$, that is, $F^*$: 
\begin{eqnarray}
F^* &:=& \nabla_\theta f^T \nabla_\theta f/T.  
\end{eqnarray}
Note that $F$ is a $P \times P$ matrix while $F^*$ is a $T \times T$ matrix. 
We can easily confirm that these $F$ and $F^*$ have the same non-zero eigenvalues.

The squared sum of the eigenvalues is given by
 $\sum_i \lambda_i^2 = \mathrm{Trace}(F^*(F^*)^T)= \sum_{st}(F_{st}^*)^2$. By using the Frobenius norm $||A||_F :=\sqrt{\sum_{ij}A_{ij}^2}$, this is  $\sum_i \lambda_i^2=||F^*||_F^2$.
 Similar to $m_\lambda$, the bias entries in $F^*$ are negligible because the number of the entries is much less than that of weight entries. Therefore, we only need to consider the weight entries. 
The $st$-th entry of $F^*$ is given by 
\begin{align}
F^*_{st} &=  \sum_l \sum_{ij}   \nabla_{W_{ij}^l}f(s)\nabla_{W_{ij}^l}f(t)/T \\ &= \sum_l  M_{l-1} \tilde{Z}^l(s,t) \hat{Z}^{l-1}(s,t)/T,
\end{align}
where we defined
\begin{equation}
\hat{Z}^l(s,t) := \frac{1}{M_l}\sum_{j}  h_j^{l} (s) h_{j}^{l}(t), \ \   \tilde{Z}^l(s,t) := \sum_{i}  \delta_i^{l} (s) \delta_{i}^{l}(t). \label{th2_62}
\end{equation}

We can apply the central limit theorem to  $\hat{Z}^{l-1}(s,t)$ and $\tilde{Z}^l(s,t)$ independently because they do not share the index of the summation. 
For $s \neq t$,  we have $\hat{Z}^l = \hat{q}^l_{st}+ \mathcal{N}(0,\hat{\gamma}/M)$ and $\tilde{Z}^l = \tilde{q}^l_{st} + \mathcal{N}(0,\tilde{\gamma}/M)$ in the limit of $M \gg 1$,  where the macroscopic variables $\hat{q}^l_{st}$ and $\tilde{q}^l_{st}$ satisfy the recurrence relations (10) and (12). Note that the recurrence relation (12) requires the Assumption 1. 
$\hat{\gamma}$ and $\tilde{\gamma}$ are constants of $O(1)$.
Then, for all $s$ and $t(\neq s)$, 
\begin{align}
F^*_{st} &= \sum_l  M_{l-1} (\tilde{q}_{st}^l +O(1/\sqrt{M})) (\hat{q}_{st}^{l-1} +O(1/\sqrt{M}))/T \\
&= \alpha \kappa_2 M/T + O(\sqrt{M})/T.
\end{align}
Similarly, for $s=t$,  we have $\hat{Z}^l = \hat{q}^l+O(1/\sqrt{M})$,  $\tilde{Z}^l = \tilde{q}^l+O(1/\sqrt{M})$ and then $F^*_{ss} =  \alpha \kappa_1 M/T+ O(\sqrt{M})/T$.

Thus, under the limit of $M \gg 1$, the dual matrix is asymptotically given by
\begin{equation}
F^* =  \alpha M K/T+ O(\sqrt{M})/T, \ \ 
K:=\begin{bmatrix} \kappa_1 & \kappa_2 & \cdots & \kappa_2 \\ 
\kappa_2 & \kappa_1 & & \vdots\\ 
\vdots & & \ddots & \kappa_2 \\ 
\kappa_2 & \cdots & \kappa_2 & \kappa_1
\end{bmatrix}. \label{K_finfin}
\end{equation}
Neglecting the lower order term, we obtain
\begin{align}
s_{\lambda} &= \sum_{s,t}^T(F^*_{st})^2/P \\
&= \alpha \left ( \frac{T-1}{T} \kappa_2^2 + \frac{1}{T} \kappa_1^2 \right ).
\end{align}
Note that, when $\hat{q}^l_{st} = 0$, $\kappa_2$ becomes zero and  the lower order term may be non-negligible. 
In this exceptional case, we have $s_{\lambda} = \alpha \kappa_1^2/T + O(1/M)$, where the second term comes from the $O(\sqrt{M})/T$ term of Eq. (A.23).
Therefore, the lower order evaluation depends on the  $T/M$ ratio, although it is outside the scope of this study. 
Intuitively, the origin of $\hat{q}^l_{st} \neq 0$ is related to the offset of firing activities $h^l_i$. The condition of $\hat{q}^l_{st} \neq 0$ is satisfied when the bias terms exist or when the activation $\phi(\cdot)$ is not an odd function. In such cases, the firing activities have the offset $\mathrm{E}[h^l_i(t)] \neq 0$. Therefore, for any input samples $s$ and $t$ ($s \neq t$), we have $\sum_i h^l_i(s) h^l_i(t)/M_l = \hat{q}^l_{st} \neq 0$ and then $\kappa_2 \neq 0$ makes $s_\lambda$ of $O(1)$.

\subsection*{(ii) $C>1$ of $O(1)$}

Here, we introduce the following dual matrix $F^*$: 
\begin{eqnarray}
F^* &:=& B^TB/T, \\
B &:=& 
[\nabla_\theta f_1 \ \ \nabla_\theta f_2 \ \  \cdots \ \  \nabla_\theta f_C],
\end{eqnarray}
where $\nabla_\theta f_k$  is a $P\times T$ matrix whose columns are the gradients on each input sample, i.e.,  $\nabla_{\theta}  f_k(t)$ $(t=1,...,T)$, 
and  $B$ is a $P \times CT$ matrix.
The FIM is represented by $F= BB^T/T$. $F^*$ is a $CT \times CT$ matrix and  consists of $T \times T$ block matrices, 
\begin{equation}
       F^*(k,k') :=  \nabla_\theta f_k^T\nabla_\theta f_{k'}/T,
\end{equation}
for $k,k'=1,...,C$.

The diagonal block $F^*(k,k)$ is evaluated in the same way as the  case of $C=1$. It becomes $\alpha M K/T$  as shown in Eq. (A.23).
The non-diagonal block  $F^*(k,k')$ has the following $st$-th entries:  
\begin{align}
F^*(k,k')_{st}&=\sum_l\sum_{ij} \nabla_{W^l_{ij}} f_k^T (s) \nabla_{W^l_{ij}}  f_{k'} (t)/T \\
&= \sum_lM_{l-1}(\sum_i  \delta_{k,i}^l (s)  \delta_{k',i}^l (t)) \hat{Z}^{l-1} (s,t)/T. 
\label{A28t}
\end{align}
Under the limit of $M \gg 1$, while  $\tilde{Z}^{l} (s,t)$ becomes $\tilde{q}^l_{st}$ of  $O(1)$,   
$(\sum_i  \delta_{k,i}^l (s)  \delta_{k',i}^l (t))$ becomes zero and its lower order term of $O(1/\sqrt{M})$ appears. This is because the different outputs ($k \neq k'$) do not share the weights $W^L_{ij}$.  We have  $\sum_i  \delta_{k,i}^L (s)  \delta_{k',i}^L (t)=0$ and then obtain $\sum_i  \delta_{k,i}^l (s)  \delta_{k',i}^l (t)=0$  ($l=1,...,L-1$) through the backpropagated chain (7).   Thus, the entries  of the non-diagonal blocks (A.28) become of $O(\sqrt{M})/T$, and we have 
\begin{equation}
  F^*(k,k') =   \alpha M K/T \delta_{k,k'}+ O(\sqrt{M})/T,
\end{equation}
where $\delta_{k,k'}$ is the Kronecker delta. 

After all, we have
\begin{align}
s_\lambda &= \sum_{k,k'}^C\sum_{s,t}^T(F^*(k,k')_{st})^2/P \\
&= C \alpha \left (\frac{T-1}{T} \kappa_2^2 + \frac{1}{T} \kappa_1^2 \right) +C O(1/\sqrt{M}) + C(C-1)O(1/M),
\end{align}
where the first term comes from the diagonal blocks of $O(M)$ and the second one is their lower order term. The third term comes from the non-diagonal blocks of $O(\sqrt{M})$.
 As one can see from here, when $C=O(M)$, the thrid term becomes non-negligible. This case is examined in Section 4.3. \QEDA

\section*{A.4 Theorem 4}

\subsection*{(i) Case of $C=1$}
Because $F$ and $F^*$ have the same non-zero eigenvalues, what we should derive here is the maximum eigenvalue of $F^*$. As shown in Eq. (A.23), the leading term of $F^*$ asymptotically  becomes $\alpha MK/T$ in the limit of $M \gg 1$. 
%Note that the Assumption 1 has been used to derive Eq. (A.22).
The eigenvalues of  $\alpha MK/T$ are explicitly obtained as follows: $\lambda_{max}=  \alpha \left ( \frac{T-1}{T} \kappa_2 +\frac{1}{T}\kappa_1 \right)  M$ for an eigenvector $e=(1,...,1)$, and $\lambda_i= \alpha  (\kappa_1 -\kappa_2)M/T $  for eigenvectors $e_1 -e_i$ ($i=2,...,T$) where $e_i$ denotes a unit vector whose entries are $1$ for the $i$-th entry and $0$ otherwise. 
Thus, we obtain 
$\lambda_{max}=\alpha \left ( \frac{T-1}{T} \kappa_2 +\frac{1}{T}\kappa_1 \right)M$.

\subsection*{(ii) $C>1$ of $O(1)$}
Let us denote $F^*$ shown in Eq. (A.31) by $F^* := \bar{F}^* + R$. $\bar{F}^*$ is the leading term of $F^*$ and given by a $CT \times CT$ block diagonal matrix whose diagonal blocks are given by $\alpha MK/T$.  $R$ denotes the residual term of $O(\sqrt{M})/T$.
In general,  the maximum eigenvalue is denoted by the spectral norm $||\cdot||_2$, that is,  $\lambda_{max}=||F^*||_2$.
 Using the triangle inequality, we have
\begin{equation}
\lambda_{max} \leq ||\bar{F}^*||_2+ ||R||_2,
\end{equation}
We can obtain $||\bar{F}^*||_2 =  \alpha \left ( \frac{T-1}{T} \kappa_2 +\frac{1}{T}\kappa_1 \right)  M$  
because the maximum eigenvalues of the diagonal blocks are the same as the case of $C=1$.  Regarding $||R||_2$,  
this is bounded by $||R||_2 \leq ||R||_F = \sqrt{C^2 \sum_{st} (O(\sqrt{M})/T)^2} = O(C\sqrt{M})$. Therefore, when $C=O(1)$, we can neglect $||R||_2$ of $O(\sqrt{M})$ compared  to $||\bar{F}^*||_2$ of $O(M)$. 

On the other hand, we can also derive the lower bound of $\lambda_{max}$ as follows. 
In general, we have 
\begin{equation}
\lambda_{max} = \max_{{\bf v}; ||{\bf v}||^2=1} {\bf v}^T F^{*} {\bf v}.
\end{equation}
Then, we find
\begin{align}
\lambda_{max} &\geq {\bf v}_1^T F^{*} {\bf v}_1,  \label{eqE6}
\end{align}
where $v_1$ is a $CT$-dimensional vector whose first $T$ entries are $1/\sqrt{T}$ and the others are $0$, that is,    $v_1=(1,...,1,0,...,0)/\sqrt{T}.$ We can compute this lower bound  by taking the sum over the entries of $F^*(1,1)$, which is equal to Eq. (A.23):   
\begin{equation}
\lambda_{max} \geq \left( \frac{T-1}{T}  \kappa_2 + \frac{1}{T} \kappa_1\right )M. 
\end{equation}
Finally, we find that the upper bound (A.34) and lower bound (A.37) asymptotically take the same value of $O(M)$, that is,  
$\lambda_{max} = \left( \frac{T-1}{T}  \kappa_2 + \frac{1}{T} \kappa_1\right )M$.

\QEDA

\section*{A.5 Case of $C=O(M)$}

The mean of eigenvalues $m_{\lambda}'$ is derived in the same way as shown in Section A.1 (ii), that is, $m'_\lambda=C\kappa_1/M$. 

Regarding the second moment $s_{\lambda}'$, the lower order term becomes non-negligible as remarked in Eq. (A.33). 
We evaluate this $s_{\lambda}'$ by using inequalities as follows: 
\begin{align}
s_{\lambda}' &= ||F^*||_F^2/P \\
&= \left (\sum_k^C || \nabla_\theta f_k^T \nabla_\theta f_{k} ||_F^2 + \sum_{k,k'}^C || \nabla_{\theta} f_k^T \nabla_{\theta} f_{k'} ||_F^2 \right)/P\\ 
& \geq   \sum_k^C || \nabla_\theta f_k^T \nabla_\theta f_{k}||_F^2  /P.
 \end{align}
 As shown in Section A.3,  for any $k$, we obtain $ || \nabla_\theta f_k^T (s) \nabla_\theta f_{k} (t)||_F^2  /P =  \alpha \left ( \frac{T-1}{T} \kappa_2^2 + \frac{1}{T} \kappa_1^2 \right )$ in the limit of $M \gg 1$.
Thus, the lower bound becomes the same form as $s_\lambda$, That is, $s_\lambda =C \alpha ( \frac{T-1}{T} \kappa_2^2 + \frac{1}{T} \kappa_1^2)$ .
In contrast, the upper bound is given by 
\begin{align}
s_{\lambda}' &=  || F||_F^2/P \\
&=  ||\sum_{k}^C  F_k  ||_F^2/P \\
&\leq   (\sum_k^C || F_k ||_F)^2/P, 
\end{align}
 where $F_k$ denotes the FIM of the $k$-th output, i.e., $F_k:=\sum_t \nabla_\theta f_k(t) \nabla_\theta f_k(t)^T/T$. Therefore,  the upper
bound is reduced to the summation over $s_\lambda$ of $C=1$. In the limit of $M \gg 1$, we obtain $s_{\lambda}' \leq C^2  || F_k ||_F^2/P =  C^2 \alpha \left (\frac{T-1}{T} \kappa_2^2 + \frac{1}{T} \kappa_1^2 \right)=Cs_\lambda$.

Next, we show inequalities for $\lambda_{max}$. We have already derived the lower bound (A.37) and this bound holds in the case of $C=O(M)$ as well. In contrast, the upper bound (A.34) may become loose when $C$ is larger than $O(1)$ because of the residual term $||R||_2$. Although it is hard to explicitly obtain the value of  $||R||_2$, the following upper bound holds and is easy to compute by using $s_\lambda$ of Eq. (14). 
Because the FIM is a positive semi-definite matrix, $\lambda_i \geq 0$ holds by definition.
Then, we have $ \lambda_{max} \leq \sqrt{\sum_i \lambda_i^2} $.   Combining this with $ s_\lambda' = \sum_i \lambda_i^2/P$,  we have $ \lambda_{max}\leq \sqrt{\alpha s_\lambda'} M \leq  \sqrt{\alpha C s_\lambda} M$.  

\QEDA

\section*{A.6 Theorem 5}

The Fisher-Rao norm is written as
\begin{align}
||\theta||_{FR} &=  \sum_{l,ij} \sum_{l',ab} F_{(l,ij),(l',ab)}   W^l_{ij}  W^{l'}_{ab},
\end{align}
where $F_{(l,ij),(l',ab)}$ represents an entry of the FIM, that is, $\sum_k^C \sum_t \nabla_{W^l_{ij}} f_k(t)  \nabla_{W^{l'}_{ab}} f_k (t)/T$.  Because $F_{(l,ij),(l',ab)}  $ includes the random variables $W^l_{ij}$ and $W^{l'}_{ab}$, we consider the following expansion.
Note that $W^l_{ij}$ and  $W^{l'}_{ab}$ are infinitesimals generated by Eq. $(8)$. 
Performing a Taylor expansion around $W^l_{ij} = W^{l'}_{ab}=0$, we obtain
\begin{align}
F_{(l,ij),(l',ab)} ({\theta}) = F_{(l,ij),(l',ab)}  ({\theta}^*) &+ \frac{\partial F_{(l,ij),(l',ab)}}{\partial W^l_{ij}} ({\theta}^*) W^l_{ij} + \frac{\partial F_{(l,ij),(l',ab)}}{\partial W^{l'}_{ab}} ({\theta}^*)  W^{l'}_{ab} \nonumber \\
&+ \mathrm{higher {\rm \mathchar`-} order \ \  terms},
\end{align}
where ${\theta}^*$ is the parameter set $\{W^l_{ij},b^l_i \}$ with $ W^l_{ij}=  W^{l'}_{ab}=0$. By substituting the above expansion into the Fisher-Rao norm and taking the average $\langle \cdot \rangle_\theta$, we obtain the following leading term:
\begin{align}
\langle F_{(l,ij),(l',ab)}   W^l_{ij}  W^{l'}_{ab}\rangle_\theta &= \langle F_{(l,ij),(l',ab)} ({\theta}^*)  W^l_{ij}  W^{l'}_{ab} \rangle_\theta \\
&=  \langle F_{(l,ij),(l',ab)} ({\theta}^*) \rangle_{\theta^*} \langle  W^l_{ij}  W^{l'}_{ab} \rangle_{\{ W^l_{ij},W^{l'}_{ab}\}} 
\end{align}
For, $(l,ij) \neq (l',ab)$, the last line becomes zero because of $\langle  W^l_{ij}  W^{l'}_{ab} \rangle_{\{ W^l_{ij},W^{l'}_{ab}\}}= \langle  W^l_{ij} \rangle_{W^{l}_{ij}} \langle W^{l'}_{ab} \rangle_{W^{l'}_{ab}}=0$. For $(l,ij) = (l',ab)$, we have $\langle  (W^l_{ij})^2 \rangle_{\{ W^l_{ij}\}}= \sigma^2_w/M_{l-1}$. After all, in the limit of $M \gg1$, we obtain 
\begin{align}
  \langle ||\theta||_{FR} \rangle_{\theta}  &=   \sum_k^C  \frac{\sum_t}{T} \sum_l \langle \sum_i  \delta^l_{k,i}(t)^2  \sum_j  h^{l-1}_j(t)^2 \rangle_{\theta^*} \frac{\sigma_w^2}{M_{l-1}} \\ 
   &= \sum_k^C  \frac{\sum_t}{T} \sigma_w^2  \sum_l \langle \tilde{q}^l \rangle_{\theta} \langle \hat{q}^{l-1} \rangle_{\theta} \\
   &=  \sigma_w^2 C \sum_l \tilde{q}^l \hat{q}^{l-1}, 
\end{align}
where the derivation of the macroscopic variables is similar to that of $m_\lambda$, as shown in Section A.1.   
Since we have $\kappa_1=  \sum_{l} \frac{\alpha_{l-1}}{\alpha} \tilde{q}^l  \hat{q}^{l-1}$, it is easy to confirm  $\langle ||\theta||_{FR} \rangle_{\theta} \leq \sigma_w^2 \alpha/\alpha_{min} C\kappa_1$. When all $\alpha_l$ take the same value, we have $\alpha/\alpha_{min}=L-1$ and the  equality holds.
\QEDA

\section*{A.7 Lemma 6}
Suppose a perturbation around the global minimum: $\theta_t = \theta^* +\Delta_t$. 
Then, the gradient update becomes
\begin{equation}
\Delta_{t+1}  \leftarrow (I-\eta F)\Delta_t + \mu(\Delta_t-\Delta_{t-1}),
\end{equation}
where we have used $ E(\theta^*)=0$ and  $ \partial E(\theta^*) /\partial \theta=0$. 

Consider a coordinate transformation from $\Delta_t$ to $\bar{\Delta}_t$ that diagonalizes $F$. It does not change the stability of the gradients. Accordingly, we can update the $i$-th component as follows:
\begin{equation}
\bar{\Delta}_{t+1,i}  \leftarrow (1-\eta \lambda_i+\mu)\bar{\Delta}_{t,i} -\mu \bar{\Delta}_{t-1,i}.
\end{equation}
Solving its characteristic equation, we obtain the general solution, 
\begin{equation}
\bar{\Delta}_{t,i} = A\lambda_+^t +B \lambda_-^t, \ \  \lambda_{\pm} =(1-\eta \lambda_i +\mu \pm \sqrt{(1-\eta\lambda_i+\mu)^2-4\mu})/2, 
\end{equation}
where $A$ and $B$ are constants. 
This recurrence relation converges if and only if $\eta\lambda_i  < 2(1+\mu)$ for all $i$. Therefore, 
$\eta< 2(1+\mu)/ \lambda_{max}$ is necessary for the steepest gradient to converge to $\theta^*$.  \ \ \ \  \QEDA

%%%%%%%%%%%%%%%%%%%%%%%%%%%%%%%%%%%%%%%%%%%%%%%%%%%%%%%%%%%%%%
\section{Analytical recurrence relations}
\label{A_thm_solvable}
\renewcommand{\theequation}{B.\arabic{equation} }
\setcounter{equation}{0}

\subsection{Erf networks}
Consider the following error function as an activation function $\phi(x)$:
\begin{equation}
\mathrm{erf}(x) = \frac{2}{\sqrt{\pi}} \int_0^x  \exp (-t^2) dt.
\end{equation}
The error function well approximates the tanh function and has a sigmoid-like shape. 
For a network with $\phi(x)=\mathrm{erf}(x)$, the recurrence relations for macroscopic variables do not require numerical integrations.

{\bf (i) $ \hat{q}^{l}$ and $\tilde{q}^l$:}
Note that  we can analytically integrate the error functions over a Gaussian distribution:
\begin{equation}
\int^{\infty}_0 Dx \mathrm{erf}(ax) \mathrm{erf}(bx) = \frac{1}{\pi} \tan^{-1} \frac{\sqrt{2}ab}{\sqrt{a^2+b^2+1/2}}.
\end{equation}
Hence, the recurrence relations for the feedforward signals (9) have the following analytical forms:
\begin{equation}
\hat{q}^{l+1} = \frac{2}{\pi}\tan^{-1} \left( \frac{q^{l+1}}{\sqrt{q^{l+1}+1/4}} \right), \ \  {q}^{l+1} = \sigma_w^2 \hat{q}^l +\sigma_b^2. 
\end{equation}

Because the derivative of the error function is Gaussian, we can also easily integrate $\phi'(x)$ over the Gaussian distribution and obtain the following analytical representations of the recurrence relations (11):
\begin{align}
\tilde{q}^l = \frac{2\tilde{q}^{l+1}\sigma_w^2}{\pi \sqrt{q^l+1/4}}, \ \ \tilde{q}^L = 1.
\end{align}

{\bf  (ii)  $ \hat{q}^{l}_{st}$ and $\tilde{q}^l_{st}$:}

To compute the recurrence relations for the feedforward correlations (10),
note that we can generally transform $I_{\phi}[a,b]$ into
\begin{equation}
I_{\phi}[a,b]= \int Dy \left( \int Dx \phi(\sqrt{a-b}x+\sqrt{b}y) \right)^2.
\end{equation}
For the error function, 
\begin{align}
\int Dx \phi(\sqrt{a-b}x+\sqrt{b}y) &= \mathrm{erf} \frac{\sqrt{b}y}{\sqrt{1+2a-2b}}, 
\end{align}
and we obtain 
\begin{equation}
I_{\phi}[a,b] =  \frac{2}{\pi} \tan^{-1} \frac{2b}{\sqrt{(1+2a)^2-(2b)^2}}.
\label{eq_82}
\end{equation}
This is the analytical form of the recurrence relation for $\hat{q}^{l}_{st}$.

Finally, because the derivative of the error function is Gaussian, we can also easily obtain 
\begin{equation}
I_{\phi'}[a,b] =  \frac{4}{ \pi \sqrt{(1+2a)^2-(2b)^2}}.
\label{eq_83}
\end{equation}
This is the analytical forms of the recurrence relations for $\tilde{q}^{l}_{st}$.

\newpage
\subsection{ReLU networks}
We define a ReLU activation as $\phi(x)= 0 \ \ (x<0), \ \  x \ \ (0 \leq x)$. 
For a network with this ReLU activation function, the recurrence relations for the macroscopic variables require no numerical integrations. 

{\bf (i) $ \hat{q}^{l}$ and $\tilde{q}^l$:}
We can explicitly perform the integrations in the recurrence relations (9) and (11): 
\begin{eqnarray}
\hat{q}^{l+1} &=& \hat{q}^{l}\sigma_w^2/2 + \sigma_b^2/2, \\
\tilde{q}^l &=& \tilde{q}^{l+1} \sigma_w^2 /2, \ \  \tilde{q}^L = 1.  
\end{eqnarray}
% Furthermore, we can obtain their analytical solutions: 
% \begin{eqnarray}
% \hat{q}^l &=& \left(1-\frac{\sigma_b^2/2}{1-\sigma_w^2/2} \right)(\sigma_w^2/2)^{l} + \frac{\sigma_b^2/2}{1-\sigma_w^2/2} , \\ 
% \tilde{q}^l &=& (\sigma_w^2 /2)^{L-l}/2.  
% \end{eqnarray}

{\bf (ii) $ \hat{q}^{l}_{st}$ and $\tilde{q}^l_{st}$:}
We can explicitly perform the integrations in the recurrence relations (10) and (12):
\begin{eqnarray}
I_{\phi}[a,b] &=& \frac{a}{2\pi} (\sqrt{1-c^2}+ c\pi/2 +c \sin^{-1} c), \\
I_{\phi'}[a,b]  &=& \frac{1}{2 \pi} (\pi/2+\sin^{-1}c),
\end{eqnarray}
where $c=b/a$.

\subsection{Linear networks}
We define a linear activation as $\phi(x)= x$. 
For a network with this linear activation function, the recurrence relations for the macroscopic variables do not require numerical integrations.

{\bf (i) $ \hat{q}^{l}$ and $\tilde{q}^l$:}
We can explicitly perform the integrations in the recurrence relations (9) and (11): 
\begin{eqnarray}
\hat{q}^l &=& \hat{q}^{l-1}\sigma_w^2 + \sigma_b^2, \label{eq_90} \\
\tilde{q}^l &=& \tilde{q}^{l+1} \sigma_w^2 , \ \ \tilde{q}^L = 1. \label{eq_91}  
\end{eqnarray}

{\bf (ii) $ \hat{q}^{l}_{st}$ and $\tilde{q}^l_{st}$:}
We can explicitly perform the integrations in the recurrence relations (10) and (12): 
\begin{eqnarray}
\hat{q}^{l+1}_{st} &=& \hat{q}^l_{st} \sigma^2_w  +\sigma_b^2, \\
\tilde{q}^l_{st} &=& \tilde{q}^{l+1}_{st} \sigma_w^2, \ \  \tilde{q}^L_{st} = 1. 
\end{eqnarray}

\vspace{50pt}

\section{Additional Experiments}
\setcounter{figure}{0}
\renewcommand{\thefigure}{\Alph{section}.\arabic{figure}}
\subsection{Dependence on $T$}

	\vskip -0.1in

\begin{figure}[H]
%	\vskip 0.2in
	\begin{center}
		\centerline{\includegraphics[width=0.62\columnwidth]{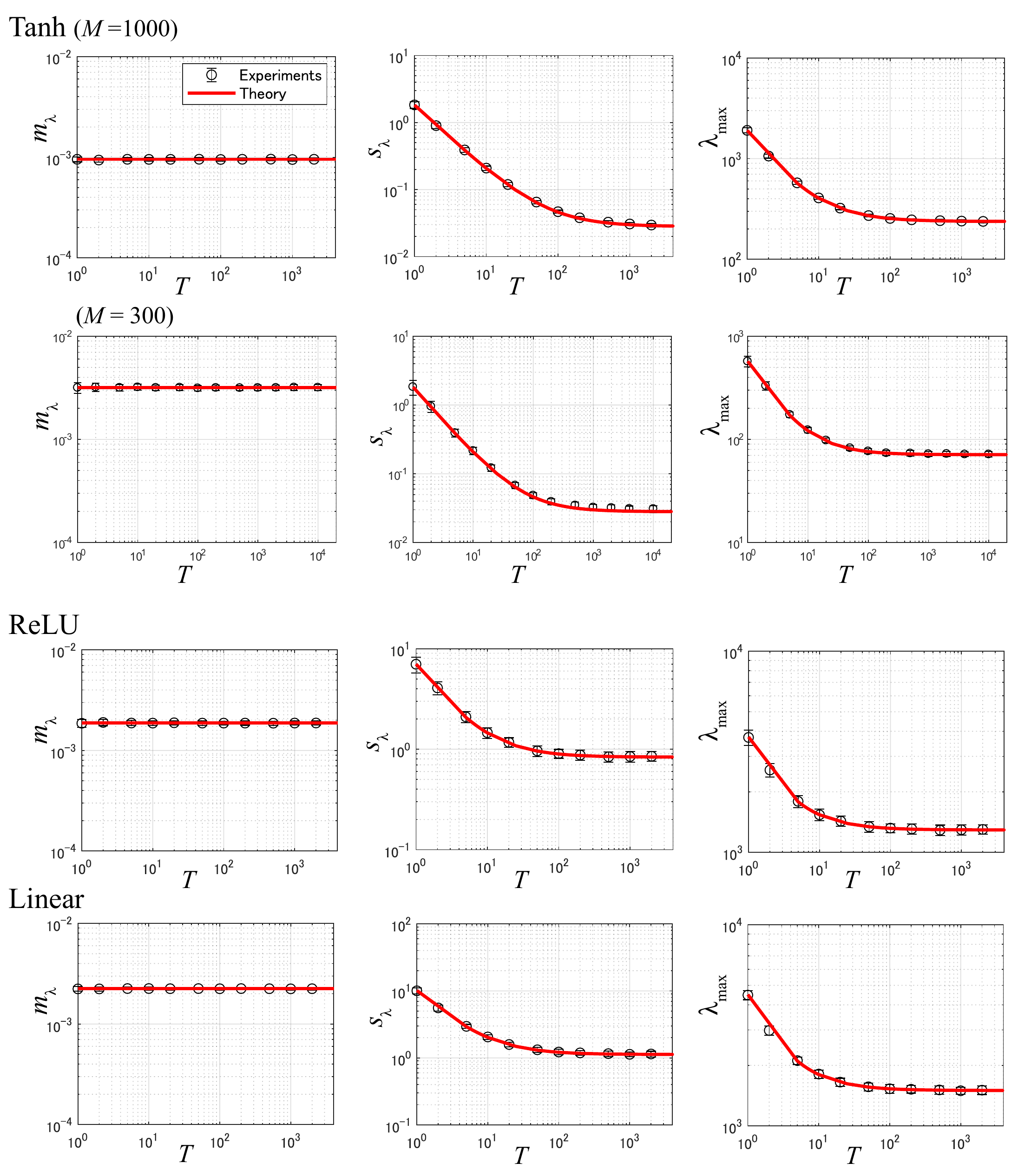}}
		\caption{Statistics of FIM eigenvalues with fixed $M$ and changing $T$  ($L=3,\alpha_l =C=1$). The red line represents theoretical results obtained in the limit of $M\gg1$.
		The first row shows results of Tanh networks with $M=1000$. The second row shows those with a relatively small  width ($M=300$) and higher $T$. We set $M=1000$ in ReLU and linear networks. The other settings are the same as in Fig. 1.  }
		\label{figA1}
	\end{center}
	\vskip -0.2in
\end{figure}

\subsection{Training on CIFAR-10}

\begin{figure}[H]
%	\vskip 0.2in
	\begin{center}
		\centerline{\includegraphics[width=0.85\columnwidth]{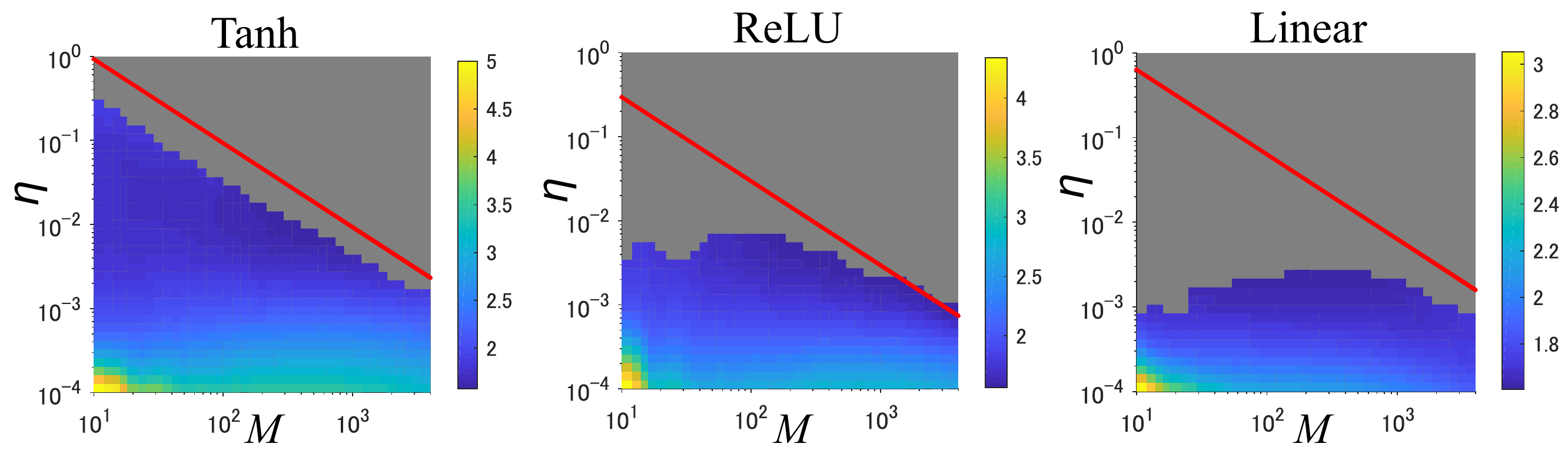}}
		\caption{Color map of training losses after one epoch of SGD training:  Tanh,  ReLU, and linear networks trained on CIFAR-10.}
		\label{figA2}
	\end{center}
	\vskip -0.2in
\end{figure}

\end{document}